\documentclass{article}

    \PassOptionsToPackage{numbers, compress}{natbib}
\usepackage[preprint]{neurips_2024}

\usepackage[utf8]{inputenc} % allow utf-8 input
\usepackage[T1]{fontenc}    % use 8-bit T1 fonts
\usepackage{hyperref}       % hyperlinks
\usepackage{url}            % simple URL typesetting
\usepackage{booktabs}       % professional-quality tables
\usepackage{amsfonts}       % blackboard math symbols
\usepackage{nicefrac}       % compact symbols for 1/2, etc.
\usepackage{microtype}      % microtypography
\usepackage{xcolor}         % colors

\usepackage[normalem]{ulem}
\usepackage[caption=true,font=footnotesize]{subfig}
\usepackage{algorithmic,amsfonts,amsmath,amssymb,booktabs,color,epsfig,float,graphicx,latexsym,makecell,multirow,palatino,soul,threeparttable}

\title{Endowing Interpretability for Neural Cognitive Diagnosis by Efficient Kolmogorov-Arnold Networks}

\author{Shangshang Yang$^{1}$ \And  Linrui Qin$^{2}$ \And Xiaoshan Yu$^{1}$ \\
$^1$School of Artificial Intelligence, Anhui University, \\
$^2$School of Computer Science and Technology, Anhui University\\
Hefei 230601, China \\
\texttt{\{yangshang0308,\ stonewallqr,\ yxsleo\}@gmail.com}
}

\begin{document}

\maketitle

\begin{abstract}
In the realm of intelligent education, cognitive diagnosis plays a crucial role in subsequent recommendation tasks attributed to the revealed students' proficiency in knowledge concepts. Although neural network-based neural cognitive diagnosis models (CDMs) have exhibited significantly better performance than traditional models, neural cognitive diagnosis is criticized for the poor model interpretability due to the multi-layer perception (MLP) employed, even with the monotonicity assumption. Therefore, this paper proposes to empower the interpretability of neural cognitive diagnosis models through efficient kolmogorov-arnold networks (KANs), named KAN2CD, where KANs are designed to enhance interpretability in two manners. Specifically, in the first manner, KANs are directly used to replace the used MLPs in existing neural CDMs; while in the second manner, the student embedding, exercise embedding, and concept embedding are directly processed by several KANs, and then their outputs are further combined and learned in a unified KAN to get final predictions. 
To overcome the problem of training  KANs slowly, we modify the implementation of original KANs to accelerate the training. Experiments on four real-world datasets show that the proposed KA2NCD   exhibits better performance than traditional CDMs, and the proposed KA2NCD still has a bit of performance leading even over the existing neural CDMs.
More importantly, the learned structures of KANs enable the proposed KA2NCD to hold as good interpretability as traditional CDMs, which is superior to existing neural CDMs. Besides, the training cost of the proposed KA2NCD is competitive to existing models.
\end{abstract}

\section{Introduction}

\begin{figure}[t]
	\centering
	\includegraphics[width=0.65\linewidth]{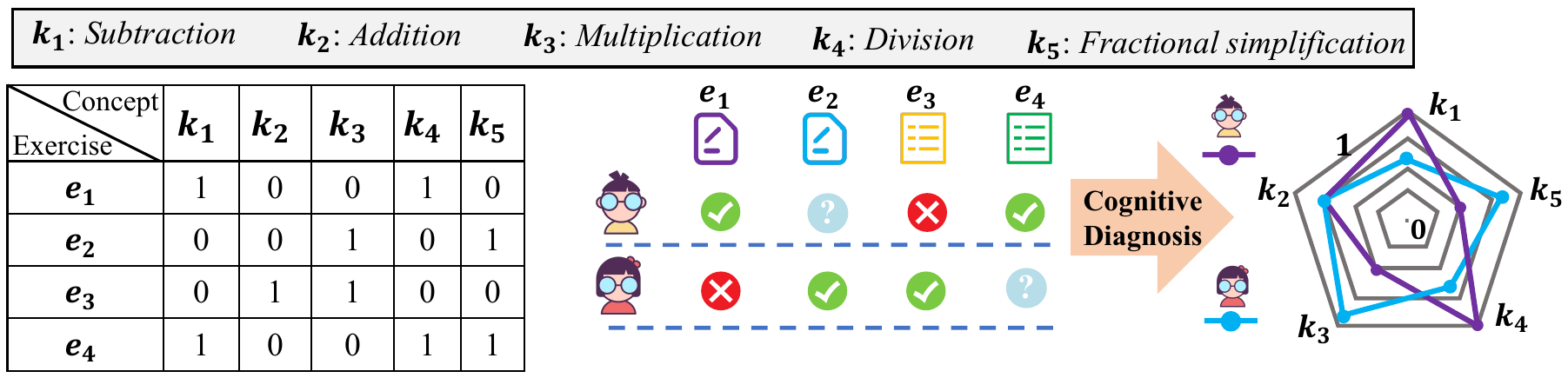}
	\caption{Illustration of cognitive diagnosis process of two students.  %Two students' response records  and $Q$-matrix are shown on the  left, while the lower right gives  the diagnosis results.
 }
	\label{fig:cognitive}
	
\end{figure}

As a crucial technique in intelligent education, cognitive diagnosis (CD)~\cite{anderson2014engaging,burns2014intelligent} is responsible for 
revealing students' proficiency in  knowledge concepts 
by mining their historical records of answering exercises.
For better understanding, an illustrative example of the CD process is given in Fig.~\ref{fig:cognitive}.
In this case, two students'  response records  are given ((i.e., $\{e_1,e_3,e_4\}$ and $\{e_1,e_2,e_3\}$)) 
 and the relationships between exercises and knowledge concepts are also 
 depicted through an exercise-concept relational matrix ($Q$-matrix for short).
As can be seen, the CD process models  the student answering prediction to obtain 
the two students' knowledge mastery states 
by mining their response records and utilizing the  $Q$-matrix.
With the diagnosed students' knowledge proficiency,   
{many subsequent tasks~\cite{Yang2023Cognitive,beck2007difficulties} on online education platforms can provide more accurate and higher quality services,} 
such as remedial instruction,  learning path recommendations, targeted training, and exercise/course assembly.

To develop convincing cognitive diagnosis models (CDMs) for meeting the demands of fast-growing  education platforms (e.g., ASSISTments~\cite{patikorn2018assistments}, PTA~\cite{hu2023ptadisc}), 
massive efforts have been devoted by researchers in different domains, mainly from two research perspectives.
The first perspective is to design completely  interpretable   CDMs  
whose components and operations are drawn from educational psychology, 
so that the users (including students, teachers, and parents)   are able to understand how the diagnosis results are obtained, trusting the results. 
The representatives  include IRT~\cite{embretson2013item},  DINA~\cite{Torre2009DINA}, MIRT~\cite{reckase2009multidimensional}, and MF~\cite{koren2009matrix}.
The second perspective is to leverage  neural networks (NNs) to 
model the student's response prediction process,  aiming to improve the prediction accuracy and provide more accurate diagnosis results for making subsequent tasks like recommendations more convincing.
This type of CDMs is called  neural cognitive diagnosis models due to the NNs used, 
and the representatives contain NCD~\cite{wang2020neural}, KaNCD~\cite{wang2022kancd}, (RCD)~\cite{gao2021rcd}, and so on~\cite{ma2022knowledge}.

Compared to traditional CDMs in educational psychology, neural CDMs indeed exhibit  significantly better performance~\cite{wang2020neural},  
which can make subsequent tasks benefit more.
Despite that,  the model interpretability  of neural CDMs is not as good as that of  
traditional CDMs because these models more or less employ the multi-layer perceptions (MLPs), 
i.e., multiple fully connected  (FC) layers~\cite{fan2021interpretability}.
It is difficult to interpret what roles the employed MLPs or FC layers play and how they process the inputs to get the output prediction~\cite{zhang2021survey} even equipped with the monotonicity assumption~\cite{samek2016evaluating}.
This monotonicity property can only ensure that their weights are positive and 
their outputs are monotonically increasing with the inputs~\cite{wang2020neural},
which can not make users understand how the diagnosis results are obtained.
As a result, neural CDMs' limited interpretability hampers their ability to engage users convincingly to some extent.

To this end, this paper aims to build more convincing CDMs by enhancing the interpretability of neural cognitive diagnosis models without sacrificing the accuracy of the diagnosis results.
 Thus, this paper proposes leveraging efficient Kolmogorov-Arnold networks for neural cognitive diagnosis models (KA2NCD) to empower the model's interpretability and maintain the accuracy of diagnosis.  Specifically, our main contributions are as follows:
\begin{itemize}

	\item 
	This paper is the first work to leverage  Kolmogorov-Arnold networks (KANs) to 
 empower the   interpretability of neural CDMs 
 while maintaining  high diagnosis accuracy. 
 To achieve this, we propose two manners of using KANS for neural CD.
 The first is replacing MLPs in neural CDMs by KANs to enhance their interpretability directly, 
and the second is designing a completely new aggregation framework consisting of multiple KANs without the help of any neural CDMs.

\item 	In the new aggregation framework, there exist two levels of KANs to handle input and output the prediction.
Several KANs at the lower level are  
used to directly process the student embedding,  exercise embedding, and concept embedding, respectively.
Then, a unified two-layer  KAN at the upper level is used to further combine and learn the outputs of lower-level KANs to get final predictions.
Besides, considering  high  runtime in the original implementation of KANs, 
we modified the implementation  of KANs to accelerate the training. 

  % TODO 翻译读起来有点别扭 检查一下
\item To validate the effectiveness of the proposed KA2NCD, 
 we compare it with some representative  CDMs on four popular education datasets.
Experimental results show that  
the proposed KA2NCD  achieves better performance than traditional CDMs and neural CDMs.
Besides, the learned structures of KANs in the proposed KA2NCD also demonstrate its higher interpretability than neural CDMs,  as good as traditional CDMs.
Moreover, the modified implementation of  KANs makes the training cost of the proposed KA2NCD  competitive to existing models.

\end{itemize}
%The rest of this paper is as follows.  Section II reviews existing CD approaches and  presents the motivation for this work. Section III presents the proposed approach KA2NCD. The experiments are shown in Section IV, and we give conclusions in Section V.

% 2.0
\section{Preliminaries and Related Work }

%To better illustrate the related work on cognitive diagnosis   and emphasize our motivation,this section  first introduces the cognitive diagnosis task and then reviews typical CDMs.

%This section  presents the preliminaries and  the related work about CD  and gives the motivation of this work.

% 2.1 介绍CD任务
\subsection{Preliminaries of Cognitive Diagnosis Task}

In the cognitive diagnosis task of intelligent education scenarios, we consider $N$ students, $M$ exercises, and $K$ knowledge concepts, represented by the sets $S = \{s_1, s_2, \dots, s_N\}$, $E = \{e_1, e_2, \dots, e_M\}$, and $C = \{c_1, c_2, \dots, c_K\}$ respectively. The platform employs an exercise-concept relation matrix provided by domain experts, referred to as the $Q$-matrix, denoted by $Q = (Q_{jk} \in {0,1})^{M \times K}$. Here, $Q_{jk} = 1$ implies that exercise $e_j$ involves knowledge concept $c_k$, and $Q_{jk} = 0$ indicates no relation between them.
Additionally, the platform maintains a log of students' responses to exercises, recorded in $R_{log}$. This log comprises triplets $(s_i, e_j, r_{ij})$, where $s_i \in S$, $e_j \in E$, and $r_{ij} \in \{0,1\}$. In this context, $r_{ij} = 1$ denotes a correct response by student $s_i$ to exercise $e_j$, whereas $r_{ij} = 0$ denotes an incorrect response.

By utilizing the students' response logs $R_{log}$ and the $Q$-matrix, the cognitive diagnosis task involves mining students' proficiency in knowledge concepts by developing a model, $\mathcal{F}$, to predict students' scores on exercises. The model $\mathcal{F}$ utilizes three types of input features to predict the score of student $s_i$ on exercise $e_j$: the student-related feature vector $\mathbf{h}_S \in \mathbb{R}^{1 \times D}$, the exercise-related feature vector $\mathbf{h}_E \in \mathbb{R}^{1 \times D}$, and the knowledge concept-related feature vector $\mathbf{h}_C \in \mathbb{R}^{1 \times K}$. Specifically, the embedding for students, exercises, and knowledge concepts can be obtained as follows:
\begin{equation}	\label{eq:val_embd}
	\small
	\left.
	\begin{aligned}
		\mathbf{h}_S = \mathbf{x}_i^S \times W_S,  W_S\in \mathbb{R}^{N\times D},
		\mathbf{h}_E = \mathbf{x}_j^E \times W_E,  W_E\in \mathbb{R}^{M\times D}, \\
		\mathbf{h}_C = \mathbf{x}_j^C \times W_Q,  W_Q\in \mathbb{R}^{K\times D}, \mathbf{x}_j^C = \mathbf{x}_j^E \times Q = (Q_{j1}, Q_{j2},\cdots, Q_{jK})\\
	\end{aligned}
\right.,	
\end{equation}

where $D$ is the embedding dimension (usually equal to $K$ for consistency),  
$\mathbf{x}_i^S \in \{0,1\}^{1\times N}$ is the  one-hot vector for student $s_i$,
 $\mathbf{x}_j^E \in \{0,1\}^{1\times M}$ is the  one-hot vector for exercise $e_j$,
 and $W_S$ and $W_E$ are trainable matrices in the embedding layers. 
 Then, the model $\mathcal{F}$  outputs the predicted response $ \hat{r}_{ij}$ as 
 \begin{equation}
      \hat{r}_{ij} = \mathcal{F}(\mathbf{h}_S,\mathbf{h}_E,\mathbf{h}_C),
 \end{equation}
where $\mathcal{F}(\cdot)$ is the diagnostic function to combine three types of inputs in different manners. 
Generally speaking, after training the model $\mathcal{F}$ based on students' response logs, 
each bit value of  $\mathbf{h}_S$ represents the student's proficiency in the corresponding knowledge concept.

\subsection{Related Work on Cognitive Diagnosis}\label{sec:related CD}

% 2.2 介绍现有CD模型类别

%Over the past many years, the educational psychology theory has played a dominant role in cognitive diagnosis due to its extremely high interpretability for CDMs.While in recent years,  NN-based neural CDMs have gained more attention because they can provide more accurate diagnosis results attributed to NNs' strong ability to model. 
In the following, some representatives of traditional CDMs and neural CDMs will be reviewed, 
which focuses on introducing their prediction process.

\subsubsection{Traditional CDMs}

% 1.传统方法 基于心理测量零与设计认知诊断模型  TODO

% ？ 这个标题用心理测量 还是 统计学
%\subsubsection{\textbf{CDMs in Psychometrics and Statistical Methods}}

%As introduced above, existing CD models can be categorized into two main types. The first type encompasses methods from the field of psychometrics and statistics, which are notable for their interpretability due to the practical significance of their parameters.

 As the most  typical CDM,  
 DINA~\cite{Torre2009DINA} outputs the prediction $\hat{r}_{ij}$ as 
 \begin{equation}
     \hat{r}_{ij} = g^{1-nt}(1-sl)^{nt}, \textrm{where}\ nt = \textstyle \prod_{k} \boldsymbol{\theta}_{k}^{\boldsymbol{\beta}_k},\ \boldsymbol{\beta} = \mathbf{h}_C.
 \end{equation}
 $\boldsymbol{\theta}\in \{0,1\}^{1\times K}$ and $\boldsymbol{\beta}\in \{0,1\}^{1\times K}$ are two binary latent features to indicate which concepts the student mastered and the exercise contains. $\boldsymbol{\theta}$ can be  obtained from $\mathbf{h}_S$ through a FC layer and Gumbel-Softmax~\cite{jang2016categorical}.
The guessing factor $g\in \mathbb{R}^{1}$ and slipping factor $sl\in \mathbb{R}^{1}$ refer to 
the probability of correctly answering the exercise only with guessing and 
the probability of mistakenly answering even with the corresponding concept mastered, and they are transformed from $\mathbf{h}_E$ by FC layers.
 As can be seen,
 the prediction process of DINA is interpretable, but 
 DINA suffers from poor prediction performance on large-scale data.

As another typical CDM, the prediction process of IRT~\cite{embretson2013item} can be denoted as follows:
\begin{equation}
    \begin{aligned}
       \hat{r}_{ij} = Sigmoid(a(\theta -\beta)),\ \theta \in \mathbb{R}^{1} = FC(\mathbf{h}_S), 
        \ &\beta \in \mathbb{R}^{1} = FC(\mathbf{h}_E), \ a \in \mathbb{R}^{1}  =FC(\mathbf{h}_E) \\
    \end{aligned},
\end{equation}
where  $\theta$ is obtained from $\mathbf{h}_S$ by an FC layer, denoting the student ability feature.
$\beta$ and $a$ are transformed from  $\mathbf{h}_S$ by two different FC layers, denoting the exercise difficulty and distinction features.
As can be seen, the prediction of IRT  can be easily understood and interpreted. 
However,  IRT also does not perform well on some complex datasets.

As a multidimensional variant of IRT, MIRT~\cite{reckase2009multidimensional} applies the same  logistic function  to the linear transformation of  the  student ability vector $\boldsymbol{\theta} \in \mathbb{R}^{1\times K}$, 
the exercise difficulty feature ${\beta}\in \mathbb{R}^{1}$, and the knowledge concept latent vector $\boldsymbol{\alpha}\in \mathbb{R}^{1\times K}$. 
That can be denoted as
\begin{equation}
    \begin{aligned}
      \hat{r}_{ij} = Sigmoid({\sum \boldsymbol{\alpha}\odot \boldsymbol{\theta}-\beta}),\   \boldsymbol{\theta} =\mathbf{h}_S,\ &\beta  = FC(\mathbf{h}_E), \ \boldsymbol{\alpha} = FC(\mathbf{h}_C)\\
    \end{aligned},
\end{equation}
where student ability feature $\boldsymbol{\theta}$ and knowledge concept latent feature $\boldsymbol{\alpha}$ are   multidimensional and can handle multidimensional data~\cite{cheng2019dirt}.
Therefore,  MIRT exhibits better performance than IRT  without losing interpretability.

Different from the above traditional CDMs,  MF~\cite{koren2009matrix} is originally devised for recommender systems but can be used for CD.
MF holds very high interpretability  because its prediction process~\cite{wang2020neural} is very easy as $\hat{r}_{ij} = \sum \mathbf{h}_S\odot \mathbf{h}_E$.
MF directly applies the inner product to student embedding $\mathbf{h}_S$ and exercise embedding $\mathbf{h}_E$ to compute the similarity.
Larger similarity represents a higher probability of correctly the student answering the exercise.
MF is quite simple yet effective compared to above CDMs.

 \subsubsection{\textbf{Neural CDMs}}
 
 % TODO 这一段的开头是否存疑

To  improve the diagnosis accuracy, 
Wang \emph{et al.} tried to incorporate NNs with high-interpretability traditional CDMs like IRT and thus proposed a neural cognitive diagnosis framework (NCD)~\cite{wang2020neural}. 
NCD's prediction is as
\begin{equation}
    \begin{aligned}
    \mathbf{f}_S = Sigmoid(\mathbf{h}_S), \ &\mathbf{f}_{diff} = Sigmoid(\mathbf{h}_E),\ f_{disc} = Sigmoid(FC(\mathbf{h}_E))  	\\
    &\hat{r}_{ij} = FC_3(FC_2(FC_1(\mathbf{y}))),\ \mathbf{y} = \mathbf{h}_C\odot(\mathbf{f}_S-\mathbf{f}_{diff} )\times f_{disc}\\
\end{aligned}. 
\end{equation}
$\mathbf{f}_S\in \mathbb{R}^{1\times K}$ represents student ability vector,
$\mathbf{f}_{diff}\in \mathbb{R}^{1\times K} $ and $f_{disc}\in \mathbb{R}^{1}$ represent exercise difficulty vector and distinction feature.
The computation process of $\mathbf{y}$ is similar to IRT, 
and  $\hat{r}_{ij}$ is obtained by applying three FC layers to $\mathbf{y}$.
It can be seen the three-FC-layer is difficult to interpret in NCD.
Although NCD attempts to utilize the monotonicity assumption to enhance its interpretability, 
the assumption can not make users understand how the diagnosis results are obtained.
In summary, NCD indeed achieves promising improvements in diagnosis accuracy 
yet sacrifices its model interpretability to some extent.

As the follow-up work of NCD, KSCD~\cite{ma2022knowledge} adjusts the positions of FC layers in NCD to obtain more meaningful student and exercise latent vectors by incorporating with knowledge concept vector first. 
Its prediction process can be denoted as
\begin{equation}
\begin{aligned}
    \hat{\mathbf{h}_S} = Sigmoid(&FC_1([\mathbf{h}_S,\mathbf{h}_C])), \hat{\mathbf{h}_E} = Sigmoid(FC_2([\mathbf{h}_E,\mathbf{h}_C]))\\
&\hat{r}_{ij} = (\sum{\mathbf{h}_C\times(\hat{\mathbf{h}_S}-\hat{\mathbf{h}_E})})/D\\
\end{aligned},
\end{equation}
where $\hat{\mathbf{h}_S}$ and $\hat{\mathbf{h}_E}$ are two latent vector with length of $D$.
Despite the promising performance of KSCD, we can see the second equation can be interpreted yet the first one containing two FC layers holds poor interpretability.

 As another representative of neural CDMs, 
 the diagnostic function of RCD~\cite{gao2021rcd} gets the prediction $\hat{r}_{ij}$ as follows:
 \begin{equation}
     \begin{aligned}
         \hat{\mathbf{h}_S} = Sigmoid(&FC_1([\mathbf{h}_S,\mathbf{h}_C])), \hat{\mathbf{h}_E} = Sigmoid(FC_2([\mathbf{h}_E,\mathbf{h}_C]))\\
         &\hat{r}_{ij} = Sigmod(\frac{1}{D}\sum FC_3(\hat{\mathbf{h}_S}-\hat{\mathbf{h}_E}) )
     \end{aligned}.
 \end{equation}
Although previous papers~\cite{ma2022prerequisite} show RCD holds dominant performance over other neural CDMs, it can be seen that RCD's interpretability is worse than  KSCD 
 because the second equation further contains an FC layer.

% In~\cite{gao2021rcd}, Gao \emph{et al.} proposed   RCD to  incorporate the model inputs with the prior relations between knowledge concepts.
% To be specific, students, exercises, and concepts are first built as a hierarchical graph.
% This graph  contains a student-exercise interaction map, a concept-exercise correlation map, 
% and a concept dependency map that is extracted from the prior relations between knowledge concepts.
% Then, a multi-level attention NN is used to achieve node aggregation of the hierarchical graph, 
% and the aggregated node features are used as three input vectors, $\mathbf{h}_S$, $\mathbf{h}_E$, and $\mathbf{h}_C$, to improve the model performance. 

% 2.3 介绍KAN   口头语描述太干了 公式 both for model and description
\subsection{Kolmogorov-Arnold Networks (KANs)}

To address the lack of interpretability in existing neural CDMs, which can be mainly attribute to the opaque nature of the MLP.
This paper aims to incorporate KANs into neural CDMs or directly leverage KANs for cognitive diagnosis because high model interpretability of KANs as shown in~\cite{liu2024kan}. 
In the following,  KANs will be introduced briefly.

A $L$-layer MLPs can be written as interleaving of  transformations ${W}$ and  activations $\sigma$:
 \begin{equation}
\operatorname{MLP}(\mathbf{x})=\left({W}_{L-1} \circ \sigma \circ {W}_{L-2} \circ \sigma \circ \cdots \circ {W}_{1} \circ \sigma \circ {W}_{0}\right) \mathbf{x},
\end{equation}
which approximates complex functional mappings through multiple layers of nonlinear transformations. However, its deeply opaque nature constrains the model's interpretability, posing challenges to intuitively understanding the internal decision-making process.

To address the issues of low parameter efficiency and poor interpretability in MLPs, Liu \emph{et al.}~\cite{liu2024kan} introduced the Kolmogorov-Arnold Network (KAN) that is inspired by Kolmogorov-Arnold representation theorem ~\cite{kolmogorov1961representation} ~\cite{braun2009constructive}. 
Similar to MLP, a $L$-layer KAN can be described as a nesting of multiple KAN layers:
\begin{equation}
    \operatorname{KAN}(\mathbf{x})=\left(\boldsymbol{\Phi}_{L-1} \circ \boldsymbol{\Phi}_{L-2} \circ \cdots \circ \boldsymbol{\Phi}_{1} \circ \boldsymbol{\Phi}_{0}\right) \mathbf{x},
\end{equation}
where $\boldsymbol{\Phi}_i$ represents the $i$-th layer of the whole KAN network. For each KAN layer with $n_{in}$ -dimensional input and $n_{out}$ -dimensional output,  $\boldsymbol{\Phi}$ consist of $ n_{in} * n_{out}$ 1-D learnable activation functions $\phi$:
\begin{equation}
    \boldsymbol{\Phi}=\left\{\phi_{q, p}\right\}, \quad p=1,2, \cdots, n_{\text {in }}, \quad q=1,2 \cdots, n_{\text {out }}.
\end{equation}

When computing the result of the KAN network from layer $l$ to layer $l+1$, it can be represented in matrix form as follows:
\begin{equation}
    \mathbf{x}_{l+1}=\underbrace{\left(\begin{array}{cccc}
\phi_{l, 1,1}(\cdot) & \phi_{l, 1,2}(\cdot) & \cdots & \phi_{l, 1, n_{l}}(\cdot) \\
\phi_{l, 2,1}(\cdot) & \phi_{l, 2,2}(\cdot) & \cdots & \phi_{l, 2, n_{l}}(\cdot) \\
\vdots & \vdots & & \vdots \\
\phi_{l, n_{l+1}, 1}(\cdot) & \phi_{l, n_{l+1}, 2}(\cdot) & \cdots & \phi_{l, n_{l+1}, n_{l}}(\cdot)
\end{array}\right)}_{\boldsymbol{\Phi}_{l}} \mathbf{x}_{l}.
\end{equation}

In conclusion, KANs differentiate themselves from traditional MLPs by using learnable activation functions on the edges and parametrized activation functions as weights, eliminating the need for linear weight matrices. This design allows KANs to achieve comparable or superior performance with smaller model sizes. Moreover, their structure enhances model interpretability without compromising performance, making them suitable for applications like scientific discovery. In cognitive diagnostic tasks, KANs 
may offer precise diagnosis and analysis of learners' knowledge structures, aiding personalized teaching and precision education with intuitive data interpretation.

\begin{figure}
    \centering
    \includegraphics[width=1\linewidth]{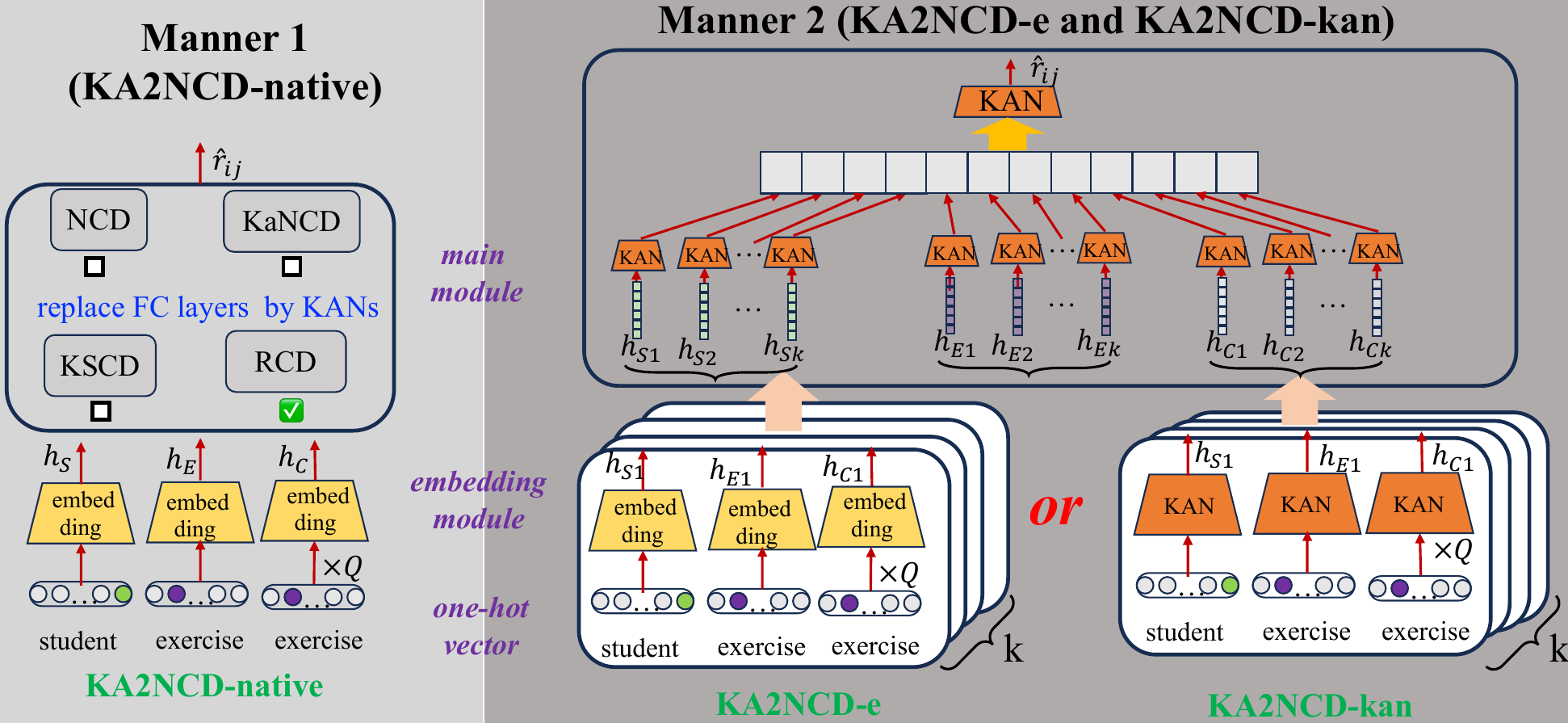}
    \caption{The Overview of the proposed KA2NCD, containing two implementation manners.}
    \label{fig:overview}
\end{figure}

\section{The Proposed KA2NCD}
% This section will first present the proposed KAN2CD framework,
% and  then sequentially

\textbf{Overview.} Figure~\ref{fig:overview} presents the overview of the proposed KA2NCD containing two implementation manners.
In the first manner (termed \textbf{KA2NCD-native}), 
all FC layers (MLPs) in the utilized CDM (NCD or KaNCD or KSCD or RCD) will be replaced 
by KANs with the same settings. 
In the second manner, there are two levels of KANs in the main module and two alternative embedding modules. 
In the main module,  KANs at the lower level process the received student, exercise, and concept embeddings, while a KAN at the upper level combines and learns the lower KANs' outputs to get the prediction.
For the second manner, \textbf{KAN2CD-e}  adopts the common embedding layers as its embedding module, while \textbf{KAN2CD-kan} adopts KANs as its embedding module.

%数据集设置 网络模型设置 先直接填进去
%实验分三块 1.主实验 4大行 DINA IRT MF MIRT传统认知诊断模型的
%  NCD KSCD KaNCD 
% 以上方法加上KAN 的效果是如何
%method2 效果如何 

% Math 09 SLP FRUSUB数据集 （小的数据集）
% 可视化实验 都跑

% 参数分析实验
% 别的数据集补

% 3.0 KAN-CD的方法

% 公式1  
% embedding增强
% F 认知诊断的诊断公式 → 包括两个组成部分
% g函数是什么
% f函数是什么 
% 介绍每个变量是什么 
% 总分   概览 每个部分包括什么东西   （如果有 就继续细分 
% overview → 每个模块的内容 

% 2种路线 1.使用KAN直接替代MLP分析KAN能否为多维的embedding产生可解释的结果

%                    不知道会不会用：  3. 寻找现有的可解释的CD模型，产生有现实意义的中间值，基于这些中间值再产生新的诊断公式
\subsection{Manner 1: KA2NCD-native}
% main idea

%method1里，你的F有两部分构成，首先是已有的CDM的公式，然后是KAN公式
%To address the issue of insufficient interpretability in existing deep learning-based CDMs,

KA2NCD-native directly replaces all FC layers in the utilized neural CDM. 
For example, when adopting the NCD as the main module, 
KA2NCD-native (can be denoted as NCD+) outputs the prediction $\hat{r}_{ij}$ as
\begin{equation}
    \begin{aligned}
    \mathbf{f}_S =& Sigmoid(\mathbf{h}_S), \ \mathbf{f}_{diff} = Sigmoid(\mathbf{h}_E),\ f_{disc} = Sigmoid(KAN_1(\mathbf{h}_E))  	\\
    &\hat{r}_{ij} = KAN_2(\mathbf{y}),\ \mathbf{y} = \mathbf{h}_C\odot(\mathbf{f}_S-\mathbf{f}_{diff} )\times f_{disc}\\
\end{aligned}.
\end{equation}
  $KAN_1(\cdot)$ is used to get the exercise difficulty scalar, and $KAN_2(\cdot)$ is used to get $\hat{r}_{ij}$ from $\mathbf{y}$. 
Due to the page limit, more KA2NCD-native variants (taking  KaNCD, KSCD, and RCD as the main modules) can be found in \textbf{Appendix},  denoted as KaNCD+, KSCD+, and RCD+.

\subsection{Manner 2: KA2NCD-e and KA2NCD-kan}
% 3.2 这个时候重新以F的视角介绍这次的F是怎么设计的 首先 q s k → embedding（采用kan） 然后 embedding → result 使用kan怎么用F表示
%仅仅使用KAN来替换MLP结构并不一定充分，我们希望使用KAN 
% 2. KAN → 多个embedding → 多个scalar → scalar自由拟合出一个合适的模型

Different from manner 1 under existing CDMs,  in manner 2, a novel aggregation framework is designed for cognitive diagnosis based on KANs, which consists of two modules, 
i.e., the embedding module to get the input embedding and the main module to process input embedding and get the prediction.

\subsubsection{The embedding module}
This paper designs two alternative embedding modules: the vanilla embedding module using the embedding layers and the KAN embedding module using KANs to get the embedding.
For whichever type of embedding module,there are $k$ sub-embedding modules to get $k$ different initial embeddings $\{\mathbf{h}_{Si},\mathbf{h}_{Ei},\mathbf{h}_{Ci}| 1\leq I \leq k\}$.
Multiple embeddings will provide diverse representations and  may cause better learning results, 
which is similar to multiple heads in Transformer~\cite{vaswani2017attention}

For the vanilla embedding module,  each sub-embedding module's forward process is the same as Eq.(\ref{eq:val_embd}), and the  process of the vanilla embedding module can be denoted as 
\begin{equation}
    \begin{aligned}
    \mathbf{h}_{Si} = \mathbf{x}_i^S &\times W_{Si},  W_{Si}\in \mathbb{R}^{N\times D},
		\mathbf{h}_{Ei} = \mathbf{x}_j^E \times W_{Ei},  W_{Ei}\in \mathbb{R}^{M\times D} \\
		&\mathbf{h}_{Ci} = \mathbf{x}_j^C \times W_{Qi},  W_{Qi}\in \mathbb{R}^{K\times D}\\
    \end{aligned}, 1\leq i\leq k.
\end{equation}

For the KAN embedding module, each sub-embedding module's forward process is as
\begin{equation}
    \begin{aligned}
    \mathbf{h}_{Si} = KAN_{1*k}(\mathbf{x}_i^S|\Phi_{1*k}), &\ 
		    \mathbf{h}_{Ei} = KAN_{2*k}(\mathbf{x}_i^E|\Phi_{2*k}), \Phi_2,\\
      &\mathbf{h}_{Ci} = KAN_{3*k}(\mathbf{x}_i^C|\Phi_3), \Phi_3,
    \end{aligned}, 1\leq i\leq k.
\end{equation}
 $\mathbf{h}_{Si}$ has a length of $D$, and thus $\Phi_{1*k}$ in $KAN_{1*k}(\cdot)$  contains $D \times N$ learnable functions  to learn the embedding $\mathbf{h}_{Si}$.
 $\Phi_{2*k}$ and $\Phi_{3*k}$ contains $D \times M$ and $D \times K$ learnable functions.  

\subsubsection{The main module}

After getting the input embedding set $H = \{H_1,\dots,H_{3*k}\} =\{\mathbf{h}_{Si},\mathbf{h}_{Ei},\mathbf{h}_{Ci}| 1\leq I \leq k\}$, the main module is used to porocess these embeddings to get the final prediction $\hat{r}_{ij}$ 
by two levels of KANs.

In the lower level, there are $3\times k$ KANs used for handling the input embedding set and obtaining
the latent vector $\mathbf{v}=\{v_1,v_2,\dots,v_{3*k}\}$, where the forward pass is as follows:
\begin{equation}
    v_i = KAN_i^{low}(H_i|\Phi_i^{low}),\ 1\leq i\leq 3*k.
\end{equation}

Afthewards, in the upper level,  a unified $2$-layer  KAN $KAN^{up}$ is further used to process to the latent vector $\mathbf{v}$ as 
\begin{equation}
    \hat{r}_{ij} = KAN^{up}(\mathbf{v}| \Phi^{up}_1,  \Phi^{up}_2) = \Phi^{up}_2 \circ \mathbf{ls} =\Phi^{up}_2 \circ\Phi^{up}_1 \circ \mathbf{v},
\end{equation}
where $\Phi^{up}_1$ contains $D \times 3*k$ learnable functions and
$\Phi^{up}_2$ contains $1\times K$ learnable functions. 
Note that there is a latent vector $\mathbf{ls}\in \mathbb{R}^{1\times K}$ with length of $K$, 
which can be used to represent the student's knowledge mastery vector.

While in manner 1, the student's knowledge mastery vector (i.e., the student knowledge proficiency vector) is still represented by the latent student ability vector. 
For example, $\mathbf{f}_S$ represents the student knowledge proficiency in NCD+ and KaNCD+,  
while $\hat{\mathbf{h}_S}$ represents the student knowledge proficiency in KSCD+ and RCD+.
\subsection{Model Training and Implementation Modification}

\textbf{Model Training.} To train the proposed KA2NCD model, the Adam optimizer~\cite{kingma2014adam}  is used to mine the  following 
cross entropy loss~\cite{de2005tutorial} between model's output  $\hat{r}_{ij}$ and  true  response $r_{ij}$:
\begin{equation}
    \mathcal{L} = - \textstyle \sum_{\left(s_{i}, e_{j}, r_{i j}\right) \in \mathcal{R}_{log}}\left(r_{i j} \log y_{i j}\right)+\left(1-r_{i j}\right) \log \left(1-y_{i j}\right).
\end{equation}
\textbf{Implementation Modification.} 
The original implementation of KANs is not efficient because all intermediate variables $X$ need to be expanded to perform different pre-given activation functions, which will require more memory and cost more to train. 
Considering most activation functions can be   linear combinations~\cite{de1978practical} of a fixed set of B-splines basis functions $B=\{B_1(\cdot),\dots,B_L(\cdot)\}$, 
the process of one activation function can be rewritten as $[B_1(X),\dots, B_L(X)]\times W_{linear}, W_{linear}\in \mathbb{R}^{L\times 1}$, i.e., activation with multiple basis functions and combine them linearly.

%loss $\mathcal{L}$.
%As a result, the mining process of student's knowledge proficiency can befinished  by modelling the student’s exercising performance prediction task. 
%To measure the difference between the  model's output  $\hat{r}_{ij}$ predicted for  student-exercise interaction  and student's true exercising response $r_{ij}$:
%the common binary cross entropy loss~\cite{de2005tutorial} as follows is used:

% 4.0实验部分
\section{Experiments}
The following experiments aim to answer the following researcher questions:
\begin{itemize}
    \item \textbf{RQ1}: How about the effectiveness of  manner 1 of KA2NCD? 
    \item \textbf{RQ2}: How about the effectiveness of manner 2 of KA2NCD, i.e., KA2NCD-e(-kan)? 
    \item  \textbf{RQ3}: How about the interpretability (visualization) of KA2NCD?
    \item  \textbf{RQ4}: How about the efficiency and complexity of the proposed KA2NCD?
\end{itemize}

\begin{table}[t]
\small
		\renewcommand{\arraystretch}{0.5}
		\caption{{Statistics of four popular education datasets.}}
			\centering
	\setlength{\tabcolsep}{0.7mm}{
		\begin{tabular}{ccccc}
			\toprule
			\textbf{Dataset}
                        & ASSISTments   ~\cite{feng2009addressing}      $^{\textcircled{1}}$
                        & SLP           ~\cite{slp2021edudata}          $^{\textcircled{2}}$
                        & JunYi         ~\cite{chang2015modeling}       $^{\textcircled{3}}$
                        & FrcSub        ~\cite{FrcSub}                  $^{\textcircled{4}}$\\
			\midrule
			\# Students/Concepts/Exercises      & 4,163/123/17,746             & 1,499/34/907     &1000/39/712       &536/8/20    \\
			%\# Concepts      & 123       & 34        &39         &8      \\
			%\# Exercises     & 17,746    & 907       &712        &20     \\
			\# Response logs & 324,572   & 57,244    &203,945    &10,720 \\
			\bottomrule
	\end{tabular}}

%	\begin{tablenotes}
		\centering
%\item $\textcircled{1}$ \url{https://sites.google.com/site/assistmentsdata/} 
%\item $\textcircled{2}$ \url{https://aic-fe.bnu.edu.cn/en/data/index.html}
%\end{tablenotes}	
	\label{tab:dataset}
\end{table}

% 4.1数据集描述+实验设置
\subsection{Experimental Settings}

% 第二块实验 方法一1和方法2学出来的结构是什么样的 画图 分析
%   https://www.latex-tables.com/#

\begin{table}[]
    \centering

    \renewcommand{\arraystretch}{0.6}
        \caption{Performance comparison on four datasets. The best, second-best, and third-best results are in bold, underlined, and underwaved:  KA2NCD-native's results are not involved.}

        \setlength{\tabcolsep}{1.1mm}{
    \begin{tabular}{c|cc|cc|cc|cc}
    \toprule
\textbf{Dataset}           & 	\multicolumn{2}{c}{\textbf{Assistments}} &          	\multicolumn{2}{c}{\textbf{SLP}}     &         	\multicolumn{2}{c}{\textbf{JunYi}}   &         	\multicolumn{2}{c}{\textbf{FrcSub}}  
\\
\hline
Method            & AUC$\uparrow$         & ACC$\uparrow$     & AUC$\uparrow$     & ACC$\uparrow$     & AUC$\uparrow$     & ACC$\uparrow$     & AUC$\uparrow$     & ACC$\uparrow$     \\
\hline

IRT               & 72.02\%     & 70.25\% & 80.91\% & 74.29\% & 74.80\% & 72.74\% & 80.63\% & 57.14\% \\
\hline
MIRT              & 65.84\%     & 63.90\% & 72.78\% & 71.90\% & 69.59\% & 69.50\% & 81.93\% & 69.12\% \\
\hline
DINA              & 72.15\%     & 68.06\% & 77.24\% & 71.43\% & 75.81\% & 68.18\% & 80.66\% & 83.12\% \\
\hline
MF                & 70.55\%     & 68.26\% & 79.22\% & 72.80\% & 79.48\% & 74.15\% & 84.10\% & 84.10\% \\
\toprule

NCD               & 74.84\%     & 72.15\% & 84.76\% & 80.72\% & 80.70\% & 76.73\% & 90.12\% & 70.15\% \\
NCD+              & 75.71\%     & 71.91\% & 84.72\% & 80.38\% & 80.38\% & 76.12\% & 90.66\% & 74.30\% \\
%Imp.(\%)          &             &          &        &          &        &           &       &           \\
\hline
KaNCD             & \uwave{76.44}\%     & \textbf{73.33}\% & 85.21\% & \underline{81.61}\% & 80.80\% & 76.15\% & 90.11\% & 76.68\% \\
KaNCD+            & 76.99\%     & 73.54\% & 85.25\% & 81.91\% & 82.06\% & 77.23\% & 91.44\% & 78.36\% \\
%Imp.(\%)          &             &          &        &          &        &           &       &           \\
\hline
RCD               & 75.91\%     & 72.99\% & \underline{85.57}\% & 79.37\% & \textbf{83.25}\% & \textbf{78.67}\% & 89.39\% & 73.83\% \\
RCD+              & 77.10\%     & 73.78\% & 86.38\% & 80.12\% & 83.33\% & 78.76\% & 89.46\% & 74.53\% \\
%Imp.(\%)          &             &          &        &          &        &           &       &           \\
\hline
KSCD              & \underline{76.55}\%     & \underline{73.04}\% & \underline{85.90}\% & \uwave{81.02}\% & \uwave{82.17}\% & \uwave{77.83}\% & 90.49\% & 80.27\% \\
KSCD+             & 76.72\%     & 73.01\% & 86.06\% & 80.87\% & 83.43\% & 78.67\% & 90.66\% & 82.93\% \\
%Imp.(\%)          &             &          &        &          &        &           &       &           \\
\toprule
KA2NCD-e           & \textbf{76.64}\%     & \uwave{72.96}\% & \textbf{86.08}\% & \textbf{82.66}\% & \underline{83.18}\% & \underline{78.39}\% & \textbf{91.27}\% & \textbf{84.58}\% \\
KA2NCD-kan           & 72.89\%     & 71.55\% & 85.91\% & 81.91\% & 83.14\% & 78.41\% & 90.38\% & 83.30\% \\
\bottomrule
    \end{tabular}}
    \label{tab:my_label}
\end{table}
% 第三块实验 efficient 时间与原版的模型差距是多大  
% 

% 参数量

\begin{table}[]
    \centering
        \caption{Model parameter and training runtime comparisons between neural CDMs and KA2NCD on SLP \& Junyi and ASSISTments \& Junyi. ('K' means kilo, 'h' represents hours).}

        \setlength{\tabcolsep}{.13mm}{
    \begin{tabular}{c|cc|cc|cc|cc|cc}
    \toprule
Param. & NCD    & NCD+  & KaNCD & KaNCD+ & RCD    & RCD+   & KSCD  & KSCD+ & KA2NCD-kan & KA2NCD-e \\
\hline
SLP    & 232K & 83K & 91K & 50K  & 279K & 115K & 94K & 81K & 4143K & 471K  \\
Junyi  & 219K & 68K & 79K & 36K  & 266K & 97K & 74K & 67401 & 3420K & 416K  \\
\bottomrule

    \end{tabular}}

            \setlength{\tabcolsep}{.13mm}{
    \begin{tabular}{c|cc|cc|cc|cc|cc}
Runtime & NCD    & NCD+  & KaNCD & KaNCD+ & RCD    & RCD+   & KSCD  & KSCD+ & KA2NCD-kan & KA2NCD-e \\
\hline
ASSIST    & 0.12h & 0.17h & 0.15h & 0.2h  & 0.16h & 0.28h & 0.46h & 1.12 & 15h & 1.2h  \\
JunYi  & 0.02h & 0.02h & 0.06h & 0.08h  & 0.03h & 0.03h & 0.08h & 0.13h & 1.3h & 0.33h  \\
    \bottomrule

    \end{tabular}}
    \label{tab:para}
\end{table}

\iffalse

\begin{table}[]
    \centering
        \caption{The parameter number comparison between neural CDMs and KA2NCD.}
    \begin{tabular}{c|cc|cc|cc|cc|cc}
    \toprule
Params & NCD    & NCD+  & KaNCD & KaNCD+ & RCD    & RCD+   & KSCD  & KSCD+ & KAN2CDk & KAN2CDe \\
\hline
SLP    & 232012 & 83256 & 91614 & 50242  & 279351 & 115295 & 94912 & 81729 & 4143200 & 471200  \\
Junyi  & 219545 & 68339 & 79039 & 36477  & 266569 & 97813  & 74336 & 67401 & 3420600 & 416040  \\
\bottomrule
    \end{tabular}

    \label{tab:para}
\end{table}

\fi

\textbf{Datasets.}
To verify the proposed KA2NCD, we conducted  experiments on four real-world education datasets, including ASSISTments~\cite{feng2009addressing}, SLP~\cite{slp2021edudata} JunYi~\cite{chang2015modeling} and FrcSub~\cite{FrcSub}. 
 Their statistics are presented in   Table~\ref{tab:dataset}, and more details can be found in  \textbf{Appendix.}

\textbf{Compared Models.}
Four traditional CDMs (including IRT, MIRT, DINA, and MF) were taken as baselines. 
Besides, four neural CDMs (including NCD, KaNCD, KSCD, and RCD) are taken as baselines to validate the effectiveness of KA2NCD-native.

\textbf{Model  Settings.}
For all models, their embedding dimension $D$ is set to the number of concepts $K$.
The hyperparameters of comparison models follow the original papers. 
For  KA2NCD, the batch size and learning rate are set to 128 and 0.002, the training epoch number is set to 20.
All models were implemented in PyTorch and executed under an Intel 13700k CPU.
Accuracy (ACC) and \textit{Area Under the ROC Curve} (AUC) were used as evaluation metrics. 
\textbf{Our source code has been uploaded as a supplementary file.}

\subsection{Effectiveness of KA2NCD-native, KA2NCD-e, and KA2NCD-kan (RQ1 \& RQ2)}
To answer RQ1 and RQ2, Table~\ref{tab:my_label} summarizes the performance of all compared CDMs on four datasets in terms of AUC and ACC values, and we can get the following observations.

Firstly, for KA2NCD-native variants, 
NCD+, KaNCD+, KSCD+, and RCD+ hold significantly better AUC and ACC performance than four traditional CDMs on three datasets (ASSISTments, SLP, and JunYi). 
On the FrcSub dataset, their AUC values are better than four traditional CDMs but 
their  ACC values are worse than  MF, whereas neural CDMs are also worse than MF regarding ACC values.
Secondly,  on ASSISTments and FrcSub datasets,
 KA2NCD-native variants  hold better AUC values than NCD, KaNCD, KSCD, and RCD, 
 where most of their ACC values are slightly better than neural CDMs.
 While for results on SLP and JunYi,  most of  KA2NCD-native variants' results are similar to NCD, KaNCD, KSCD, and RCD, but some of their results are still superior (such as RCD+ on both datasets, KaNCD+ on JunYi, and KSCD+ on Junyi).
  Thirdly, compared to existing CDMs, KA2NCD-e nearly achieves the top-best performance on all datasets, 
 while  KA2NCD-kan's performance is worse than KA2NCD-e and not very promising, especially on ASSIStments.
 However, the performance of KA2NCD-kan on the other three datasets is still promising, which is slightly worse than KSCD but better than other CDMs.

 The first two observations show the proposed KA2NCD in manner 1 (i.e., KA2NCD-native ) is effective. 
 The third one demonstrates the superiority of KA2NCD-e and  KA2NCD-kan's effectiveness, 
 further indicating leveraging KANs for CD is promising and worth exploring.

\begin{figure}
    \centering
        \includegraphics[width=1.\linewidth]{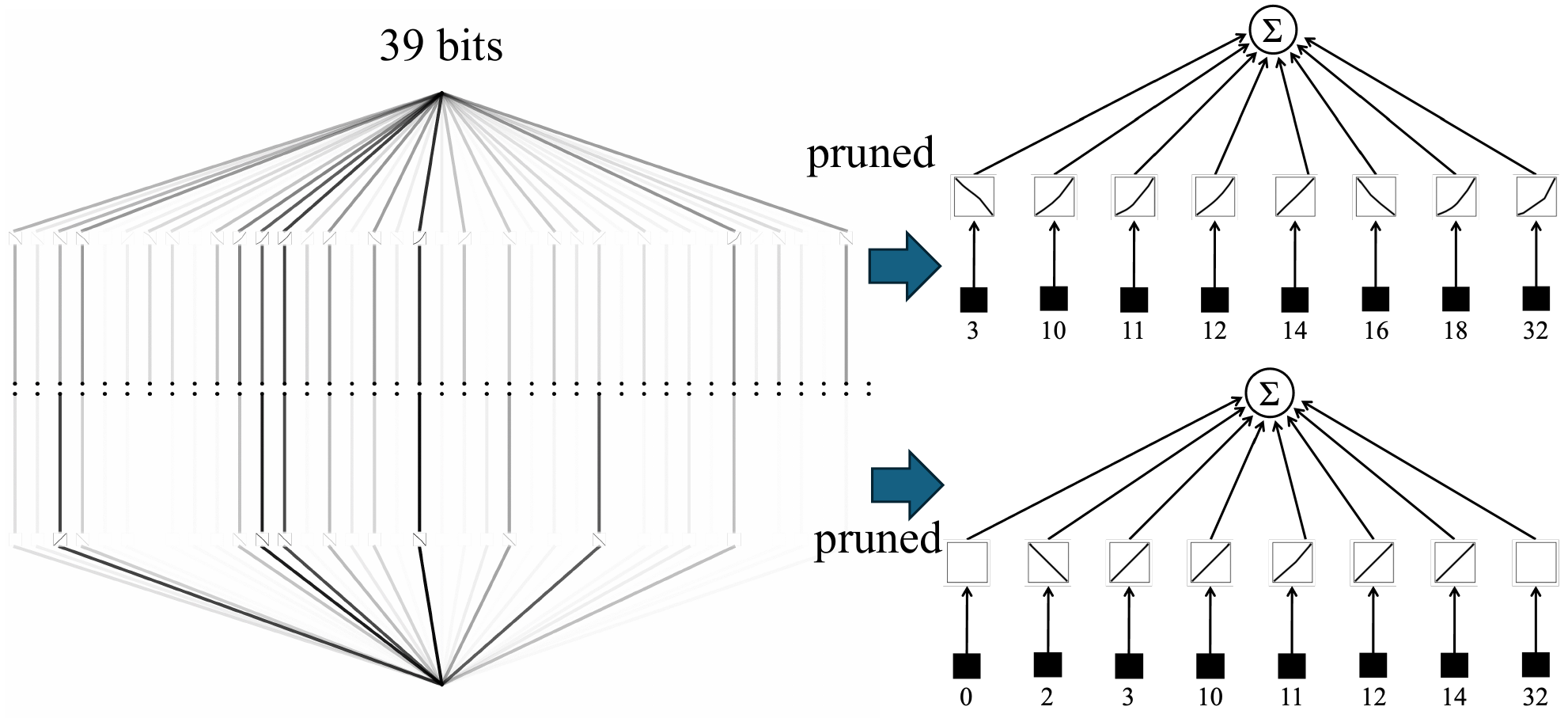}
    \caption{Visualization of NCD+ (upper)and KaNCD+ (lower) on JunYi datasets.}
    \label{fig:visualization}
\end{figure}

\subsection{Visualization of KA2NCD (RQ3)}

To answer RQ3, Figure~\ref{fig:visualization} plots the structures of NCD+ and KaNCD+ only on the JunYi dataset because its $K$ is relatively small.  The structure visualization of NCD+ and KaNCD+ on SLP can be found \textbf{in Appendix.}
In Figure~\ref{fig:visualization}, 
only the last KAN in NCD+ and KaNCD+ is plotted because they replace MLPs to get the prediction.
As can be seen, firstly, the learned structures of  their KANs are very close, which indicates the learned KAN has good transferability between two models; 
secondly,  only about 8 of 39 connections are kept in both models, which represents the learned models are easy to interpret how they work. 
 Similar observations and conclusions can be drawn from the Figures in \textbf{Appendix}, 
which shows the proposed KA2NCD can indeed enhance the interpretability of existing CDMs.

\subsection{Complexity and Efficiency of KA2NCD (RQ4)}
To show the model complexity of the proposed KA2NCD, 
Table~\ref{tab:para} presents the comparison of neural CDMs, KA2NCD-native variants, KA2NCD-e, and KA2NCD-kan regarding model parameter number and runtime of training models.
As can be seen, the parameters of KA2NCD-native are less than half of the corresponding neural CDM, 
while parameters of KA2NCD-e and KA2NCD-kan are much more than neural CDMs, especially for KA2NCD-kan.
The high parameters of KA2NCD-kan lead to its extremely high runtime.
Despite that, the runtimes of  KA2NCD-native  are close to those of neural CDMs,  
which indicates the efficiency of KA2NCD is competitive and acceptable.
The good efficiency is attributed to our modified implementations of KANs, 
because the original implementation of KANs  is very slow:   
on ASSISTments and JunYi,
NCD+ took about 0.9  and 0.3 hours, and KaNCD+ took about 1.1 and 0.35 hours, which is four times as our modified KANs.

\section{Conclusion}
In this paper, 
we proposed to enhance the model interpretability of neural CDMs by leveraging KANs, and thus we proposed KA2NCD.
There are two manners to implement KA2NCD, 
where the first manner directly replaces MLPs in existing neural CDM by KANs 
and the second manner builds a novel CDM completely with KANs.
Experimental results demonstrate the effectiveness of the proposed KA2NCD  and its high interpretability.

%Yuanchao Liu received the master degree in control theory and control engineering
%from Northeastern University, Shenyang, China, in 2019. He is
%currently pursuing the Ph.D. degree of control science and
%engineering in Northeastern University, Shenyang, China.

%His current research interests include multiobjective optimization,
%robust optimization, dynamic optimization and data-driven
%optimization.

\normalem
\bibliographystyle{plain}
\bibliography{NASCD}

\newpage
\appendix
\section{More Details}

\subsection{KA2NCD-native Variants}

\textbf{KSCD+.} When  adopting the KSCD as the main module, the prediction process of  KA2NCD-native (further denoted as KSCD+) is as  
\begin{equation}
\begin{aligned}
    \hat{\mathbf{h}_S} = Sigmoid(&KAN_1([\mathbf{h}_S,\mathbf{h}_C])), \hat{\mathbf{h}_E} = Sigmoid(KAN_2([\mathbf{h}_E,\mathbf{h}_C]))\\
&\hat{r}_{ij} = (\sum{\mathbf{h}_C\times(\hat{\mathbf{h}_S}-\hat{\mathbf{h}_E})})/D\\
\end{aligned},
\end{equation}
where the first KAN  $KAN_1(\cdot)$ is used to combine student embedding and concept embedding to get the richer student ability vector and 
the second KAN  $KAN_2(\cdot)$  is used to get the richer exercise difficulty vector.

\textbf{RCD+.} When  adopting the RCD as the main module, the prediction  of  KA2NCD-native (further denoted as RCD+) can be obtained according to the following process:
 \begin{equation}
     \begin{aligned}
         \hat{\mathbf{h}_S} = Sigmoid(&KAN_1([\mathbf{h}_S,\mathbf{h}_C])), \hat{\mathbf{h}_E} = Sigmoid(KAN_2([\mathbf{h}_E,\mathbf{h}_C]))\\
         &\hat{r}_{ij} = Sigmod(\frac{1}{D}\sum KAN_3(\hat{\mathbf{h}_S}-\hat{\mathbf{h}_E}) )
     \end{aligned}.
 \end{equation}
where the first two  KANs are the same as those of KSCD+, which is used to get a richer student ability vector and a richer exercise difficulty vector. 
The third KAN $KAN_3(\cdot)$ is used to extract the response prediction.

\textbf{KaNCD+.} When  adopting the KaNCD as the main module, we can follow the following equation to get the prediction  of  KA2NCD-native (further denoted as KaNCD+):
\begin{equation}
    \begin{aligned}
    \mathbf{f}_S =& Sigmoid(KAN_1(\mathbf{h}_S)), \ \mathbf{f}_{diff} = Sigmoid(KAN_2(\mathbf{h}_E)),\ f_{disc} = Sigmoid(KAN_3(\mathbf{h}_E))  	\\
    &\hat{r}_{ij} = KAN_4(\mathbf{y}),\ \mathbf{y} = \mathbf{h}_C\odot(\mathbf{f}_S-\mathbf{f}_{diff} )\times f_{disc}\\
\end{aligned}.
\end{equation}
It can be seen that the last two  KANs have the same effect as that of NCD+, 
and the first and second KANs are used to get the enhanced student ability vector and exercise difficulty vector.

\subsection{Details about datasets}

The detailed descriptions of four datasets are as follows:
\begin{itemize}
	\footnotesize
	\item \textbf{ASSISTments}~\cite{feng2009addressing} is an openly available dataset created in	2009 by the ASSISTments online tutoring service system. 
	Here we adopted the public corrected version that does not contain duplicate data.
	The dataset contains more than 4,000 students, nearly 18,500 exercises, and over 300,000 response logs.
	%~\cite{xiong2016going}
	
	\item  \textbf{SLP}~\cite{slp2021edudata} is another  public education dataset  published	in 2021.
	SLP collects learners' regularly captured academic performance data during their three-year studies in eight subjects: Chinese, mathematics, English, physics, chemistry, biology, history, and geography.
        % TODO 修改引用   https://pslcdatashop.web.cmu.edu/DatasetInfo?datasetId=1198
        \item  \textbf{JunYi}~\cite{chang2015modeling} JunYi is a public dataset collected by the JunYi Education Platform in Taiwan, containing nearly $20M$ responses from 1,000 students. E-Math is a private dataset collected by a well-known electronic educational product, mainly containing primary and secondary school students' math exercises and test logs.
	% The dataset contains nearly 58 thousand response logs of 1,499 students on 907 exercises.

        \item  \textbf{FrcSub}~\footnote{\url{http://staff.ustc.edu.cn/\%7Eqiliuql/data/math2015.rar}}
         The FrcSub dataset, commonly utilized for evaluating cognitive diagnosis models, focuses on fraction addition and subtraction problems. It encompasses performance data from 536 learners across 20 objective questions, with scores represented as binary values (1 for correct, 0 for incorrect). Additionally, it includes a knowledge point correlation matrix that details the relationship between the 20 questions and 8 distinct knowledge points, where a value of 1 indicates a related knowledge point, and 0 indicates no relation. This dataset serves as a valuable resource for cognitive modeling research.
\end{itemize}

For the four datasets, we filtered out students with fewer than 15 response logs to ensure sufficient data for model learning. 
Each student's logs are randomly split into training and testing datasets at the splitting ratio of 7/3.

\section{More Visualization}

\begin{figure}
    \centering
        \includegraphics[width=0.9\linewidth]{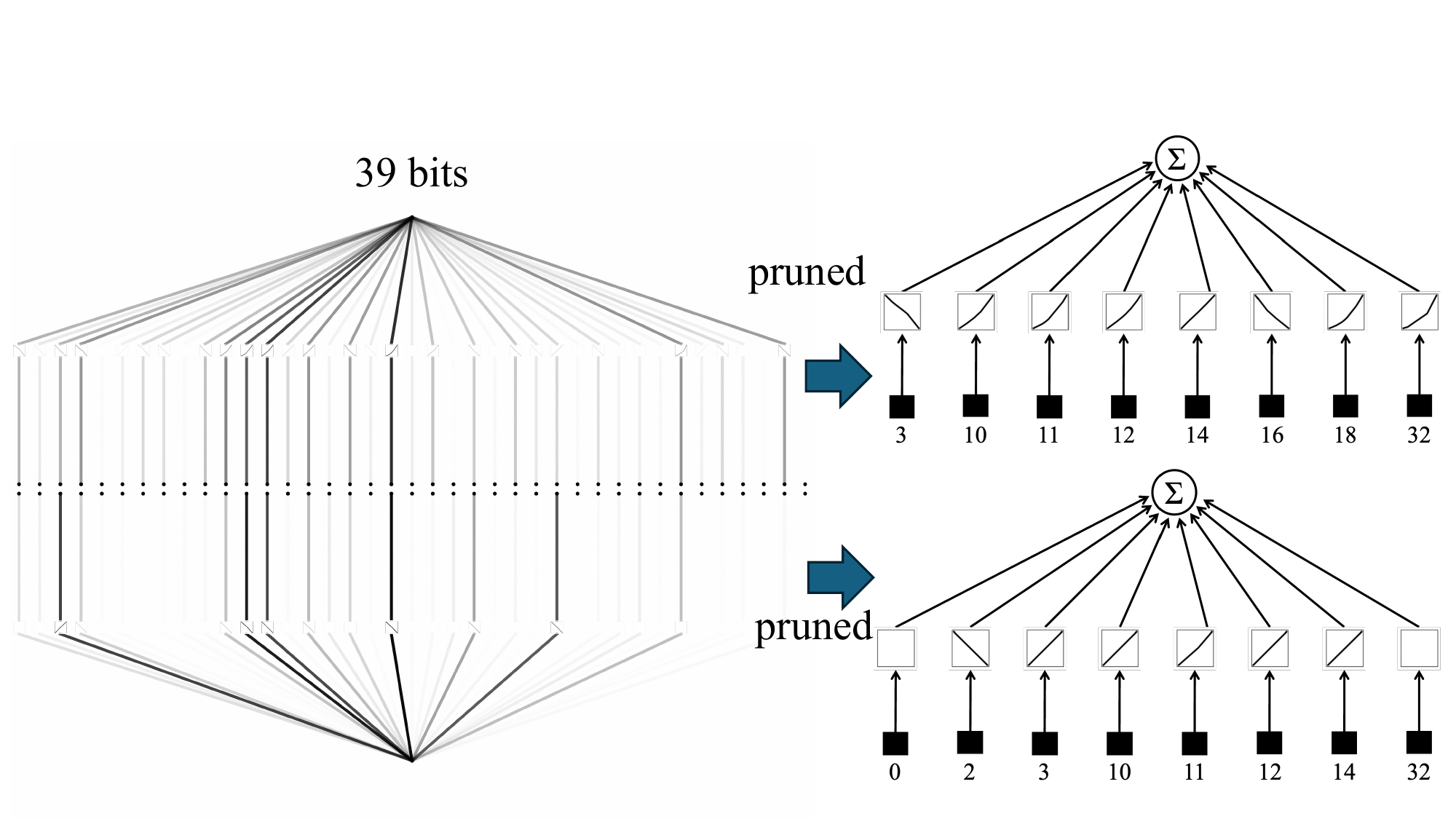}
    \caption{Overview-Visualization of KANs in NCD+ (upper)and KaNCD+ (lower) on the SLP datasets.}
    \label{fig:visualization1}
\end{figure}

\begin{figure}
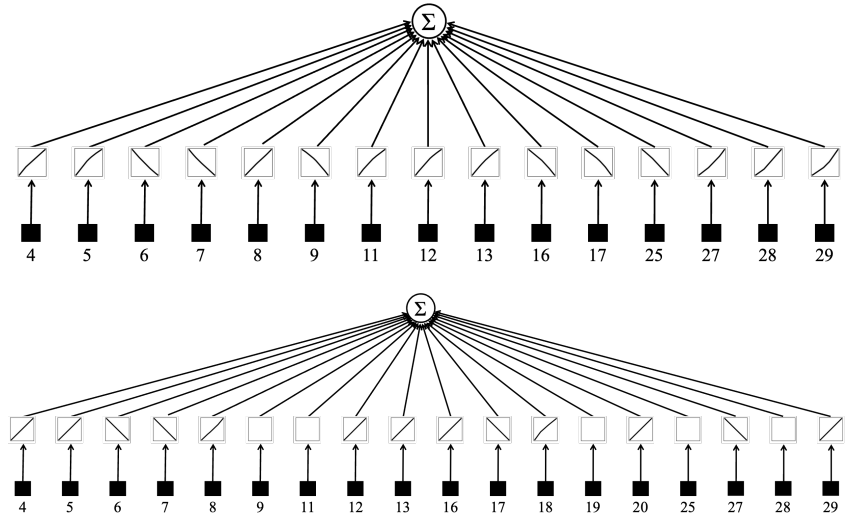

    \centering
        \includegraphics[width=0.8\linewidth]{SLP_1.pdf}
        \includegraphics[width=.8\linewidth]{SLP_2.pdf}
    \caption{ Visualization of pruned KANs in NCD+ (upper)and KaNCD+ (lower) on the SLP datasets.}
    \label{fig:visualization11}
\end{figure}
\begin{figure}
    \centering
        \includegraphics[width=0.7\linewidth]{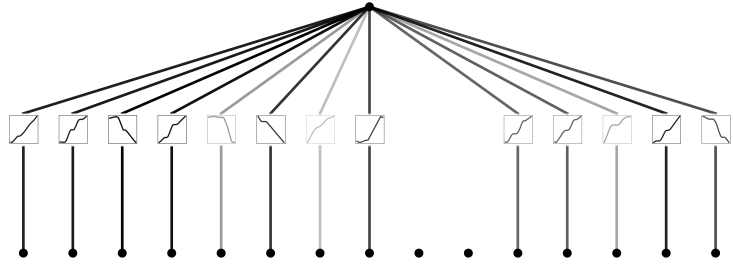}
    \caption{Visualization of the last KAN in KA2NCD-e ) on the JunYi datasets.}
    \label{fig:visualization2}
\end{figure}

\begin{figure}
    \centering
        \includegraphics[width=0.7\linewidth]{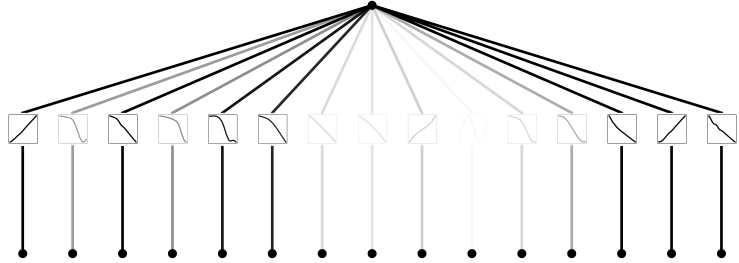}
        \caption{Visualization of KA2NCD-e (upper) on the SLP datasets.}
    \label{fig:visualization3}
\end{figure}

For deep insight into the learned KANs' structures to answer \textbf{RQ3},
Figures~\ref{fig:visualization1} and~\ref{fig:visualization11} plot the structures of NCD+ and KaNCD+ learned  on  the SLP  dataset,
Figure~\ref{fig:visualization2} plots the structures of KA2NCD-e learned on the JunYi dataset, while Figure~\ref{fig:visualization3} plots the structures of KA2NCD-e  learned on the SLP  dataset.
For easy observation, 
for Figure~\ref{fig:visualization1}, only  the last KAN in NCD+ and KaNCD+ is plotted because they replace MLPs to get the prediction;
while for Figures~\ref{fig:visualization2} and~\ref{fig:visualization3}, only the structures of the upper-level KANs are presented.

As can be seen from Figures~\ref{fig:visualization1} and~\ref{fig:visualization11},  the learned structures of   KANs are very close, which indicates the learned KAN has good transferability between two models;  about 15 of 34 connections are kept in both models, which makes the learned models easy to be interpreted. 
As can be seen from Figures~\ref{fig:visualization2} and~\ref{fig:visualization3}, 
which latent features are used and how they are combined can be obviously observed, which is very important for users to understand.

In conclusion, the proposed KA2NCD can indeed enhance the  interpretability of existing CDMs, 
and the manner 2 of the proposed KA2NCD is indeed effective,  where the learned structures are really easy to interpret.

\end{document}